\documentclass{article}
\usepackage{spconf,amsmath,graphicx,hyperref}
\usepackage[utf8]{inputenc} 
\usepackage[T1]{fontenc}    
\usepackage{url}            
\usepackage{booktabs}       
\usepackage{amsfonts}       
\usepackage{nicefrac}       
\usepackage{microtype}      
\usepackage{xcolor}         
\usepackage{graphicx} 
\usepackage{subcaption}  
\usepackage{amssymb}
\usepackage{algorithm}      
\usepackage{algpseudocode}  
\usepackage{float}      
\usepackage{caption}   
\usepackage{longtable}
\usepackage{multirow}

\title{ContraFM-S2O: Flow Matching-Based One-step SAR-to-Optical Image Translation Model with Contrastive Learning}
\name{Mingqian Yu$^{1,\#}$\thanks{\#: Equal contribution.} \qquad Wei-Kuan Chiang$^{2,\#}$  \qquad Qiurui Wang$^{3}$ \qquad Peilin Zhao$^{4,*}$\thanks{*: Corresponding author.}}
\address{$^{1}$Institute of Automation, Chinese Academy of Sciences, Beijing, China \\
        $^{2}$Department of Computer Science, The University of Manchester, Manchester, UK \\
        $^{3}$Institute of Artificial Intelligence in Sports, Capital University of Physical Education And Sports, Beijing, China \\
        $^{4}$School of Artificial Intelligence, Shanghai Jiao Tong University, Shanghai, China \\
        \texttt{yumingqian2026@ia.ac.cn}\\ 
        \texttt{chiangweikuan2004@gmail.com} \\ \texttt{wangqiurui@cupes.edu.cn} \\ \texttt{peilinzhao@sjtu.edu.cn}}
\ninept
\begin{document}
\maketitle
\begin{figure*}[t]
    \centering
    \includegraphics[width=\textwidth, height=0.45\textwidth]{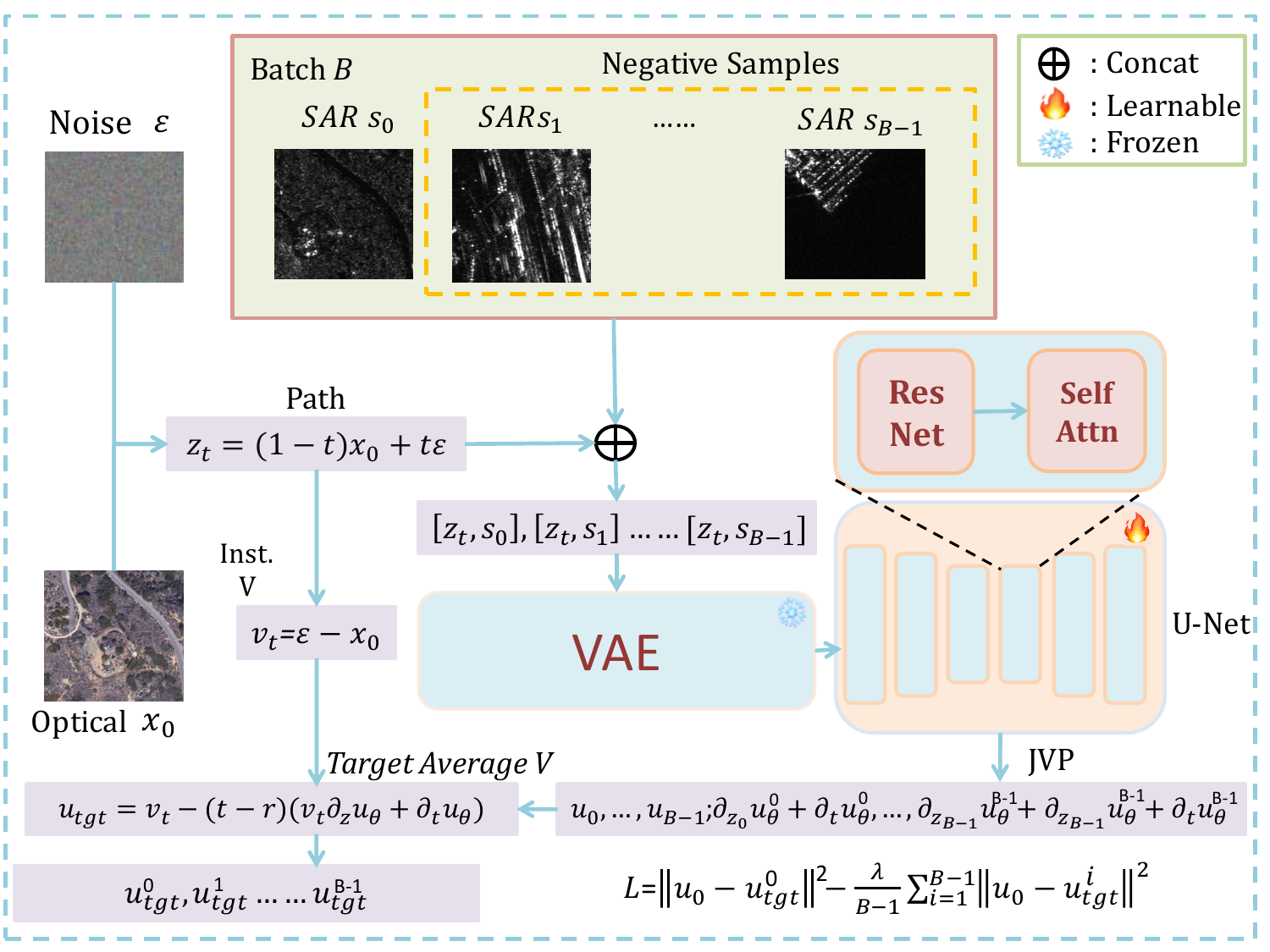}
    \caption{The training pipeline of ContraFM-S2O, which replaces instantaneous velocity with average velocity to realize one-step generation and uses contrastive learning to prevent the flows between unpaired SAR and optical images from overlapping.}
    \label{fig:ContraFM-S2O}
\end{figure*}

\begin{abstract}
In recent years, diffusion models and GAN-based models have become the mainstream approaches for SAR-to-optical image translation, owing to their advantages, such as high-quality generation and stable training. However, they have shortcomings such as high inference latency and the generated optical images suffer from low detail fidelity, often resulting in blurred edges and loss of fine textures. Thus, we propose ContraFM-S2O, which is a flow matching-based model for SAR-to-optical image translation. Unlike conventional diffusion models, ContraFM-S2O learns to predict the velocity field in training and solves ODE instead of SDE during inference to improve the sampling efficiency. In addition, ContraFM-S2O replaces instantaneous velocity with average velocity along the interpolation path to realize one-step SAR-to-optical image translation and uses contrastive learning to improve the quality of the generated optical images. Experiments show our model achieves state-of-the-art on SAR2Opt and QXS datasets, outperforming baselines, and reduces inference latency via one-step generation.
\end{abstract}
\section{Introduction}
\label{sec:intro}
SAR-to-optical image translation has emerged as a pivotal research area, aiming to leverage the all-weather, day-and-night imaging capabilities of SAR with the high interpretability and rich spectral information of optical sensors \cite{11479792}. Besides, SAR-to-optical image translation plays a critical role in emergency detection, geological monitoring and others.

GAN-based methods such as Cycle-GAN \cite{CycleGAN2017}, Pix2Pix \cite{pix2pix2017} and PCGL-GAN \cite{chen2026towards} could generate high-quality optical images but are prone to mode collapse and the generated optical images often suffer from blurred edges, distorted textures, and color discrepancies. Methods based on diffusion models like S2ODPM \cite{bai2023conditional} and cBBDM \cite{kim2025conditional} could realize stable training and generate optical images with higher detail fidelity, but the inference latency is high, which makes real-time SAR-to-optical image generation challenging. To overcome these questions, we propose ContraFM-S2O, which is capable of one-step SAR-to-optical image generation and could generate optical images with sharp edges and high detail fidelity.

To achieve one-step SAR-to-optical image generation, ContraFM-S2O replaces instantaneous velocity with average velocity, which is defined as the ratio of displacement to a time interval, with displacement given by the time integral of the instantaneous velocity \cite{geng2026mean}. Besides, to prevent the predicted average velocity fields from overlapping when different SAR images are used as conditional inputs during training, ContraFM-S2O employs contrastive learning to distinguish flows conditioned on different SAR images (see Figure \ref{fig:ContraFM-S2O}).

Our experiments validate that ContraFM-S2O could reach the sota performance across baselines and achieve 4.4× inference acceleration compared to PCGL-GAN (the second-best model). And contrastive learning contributes more to the quality of generated optical images and average velocity contributes more to inference acceleration.

\section{Related Work}
\label{sec:related}
\subsection{SAR-to-Optical Image Translation}
Recently, SAR-to-optical image translation methods have been broadly divided into those based on generative adversarial networks (GANs) and those based on diffusion models. GAN-based methods such as ICGAN \cite{yang2022sar} and PCGL-GAN \cite{chen2026towards}, are all implemented based on the discriminator-generator architecture for SAR-to-optical image translation. ICGAN enhances contour sharpness, texture fine-grainedness, and color fidelity through a parallel feature fusion generator, a multi-scale discriminator, and a chromatic aberration loss, respectively. PCGL-GAN enhances visual consistency and structural fidelity through perception correlative learning in the feature space and collaborative modeling of global-local features with cross-scale compensation. To improve the quality of the generated optical image, researchers propose methods based on diffusion models. S2ODPM \cite{bai2023conditional} uses SAR images as condition to guide the reverse denoising process, achieving superior visual quality and structural consistency compared to GAN-based methods. cBBDM \cite{kim2025conditional} incorporates explicit spatial conditions from the pixel space into the latent diffusion process to outperform GAN-based and conditional diffusion models in both structural fidelity and visual quality. However, these methods are limited by several shortcomings, including high inference latency and insufficient clarity of edge and texture details in the generated optical images and ContraFM-S2O models the average velocity to realize one-step SAR-to-optical generation and utilizes contrastive learning to distinguish transport flows between different SAR and optical images to achieve high-quality generation. 
\subsection{Flow Matching-Based Image Generation}
Flow Matching (FM) \cite{lipman2022flow} addresses image generation by learning a continuous and deterministic transport map between source and target domains through an ordinary differential equation. By directly regressing the velocity field rather than simulating a stochastic denoising process, FM achieves faster inference with minimal sampling steps. Representative works such as Rectified Flow \cite{liu2022flow}, Conditional Flow Matching \cite{tong2023conditional} and MeanFlow \cite{geng2026mean} all achieve high-quality image generation. Rectified Flow learns to transport data between two distributions along straight-line paths, enabling computationally efficient and stable image generation and translation with minimal discretization steps. Conditional Flow Matching provides a simulation-free training objective for continuous normalizing flows by conditioning on paired source and target samples, enabling generalization to arbitrary source distributions, while its optimal transport variant (OT-CFM) further improves training and inference efficiency by approximating dynamic optimal transport. MeanFlow introduces the concept of average velocity to derive a principled identity with instantaneous velocity, enabling highly effective one-step generative modeling from scratch without pre-training, distillation, or curriculum learning.

\section{Background}
\label{sec:background}
Flow Matching \cite{lipman2022flow, liu2022flow, tong2023conditional} is a family of generative models that learn to match the flows, represented by velocity fields, between two probabilistic distributions \cite{geng2026mean}. During the training phase, the basic idea of Flow Matching is to make the model's output fit the instantaneous velocity $v_t$. 
\subsection{Conditional Velocity.} Given data $x \sim p_{data}(x)$ and prior $\epsilon \sim p_{prior}(\epsilon)$, $v_t=z_t'=a_t' x+b_t' \epsilon$ ($a_t=1-t$ and $b_t=t$ commonly), where $'$ denotes the time derivative. We call $v_t$ as conditional velocity \cite{lipman2022flow}. The conditional Flow-Matching loss $\mathcal{L}_{CFM}$ is defined as follows:

\begin{equation}
    \mathcal{L}_{CFM}(\theta)=E_{t,x,\epsilon}\lVert v_\theta(z_t,t)-v_t(z_t|x) \rVert^2,
    \label{eq:1}
\end{equation}
where the target $v_t$ is the conditional velocity, since a $z_t$ can be obtained from different $(x,\epsilon)$ pairs, the same $z_t$ leads to different conditional velocities.
\subsection{Marginal Velocity.} Essentially, Flow Matching constructs a model of the expectation over every possible case, also known as the marginal velocity \cite{lipman2022flow}. 

\begin{equation}
    v(z_t,t)\triangleq E_{pt(v_t|z_t)}[v_t]
    \label{eq:2}
\end{equation}
A neural network $v_\theta$ parameterized by $\theta$ is learned to fit the marginal velocity field \cite{geng2026mean} and $\mathcal{L}_{FM}$ is defined as follows:
\begin{equation}
    \mathcal{L}_{FM}(\theta)=E_{t,p_t(z_t)}\lVert v_\theta(z_t,t)- v(z_t,t) \rVert ^{2}.
    \label{eq:3}
\end{equation} 
Although computing this loss function is infeasible due to the marginalization in Eq. \eqref{eq:1}, it is proposed to instead evaluate the conditional Flow Matching loss. Minimizing $\mathcal{L}_{CFM}$ is equal to minimizing $\mathcal{L}_{FM}$ \cite{geng2026mean,lipman2022flow}.  
\\
Given a marginal velocity field $v(z_t,t)$, samples are generated by solving an ODE for $z_t$:
\begin{equation}
    \frac{d}{dt}z_t=v(z_t,t)
    \label{eq:4}
\end{equation}
starting from $z_1=\epsilon - p_{prior}$. The solution can be written as: $z_r=z_t- \int_{r}^{t}v(z_{\tau},\tau)\, d\tau$, where we utilize $r$ to denote another time step. \cite{geng2026mean}

\section{ContraFM-S2O}
\label{sec:contrafm-s2o}

To reduce the inference latency and achieve one-step generation, ContraFM-S2O learns to predict the average velocity instead of instantaneous velocity. In addition, ContraFM-S2O uses contrastive learning to prevent the flow from overlapping and improve the quality of generated optical images.

\subsection{Average Velocity}
Motivated by MeanFlow \cite{geng2026mean}, we model the average velocity field instead of instantaneous velocity field. Firstly, we use $u$ to denote average velocity and $v$ to denote instantaneous velocity. Given two time steps $t$, $r$ the average velocity $u$ is formally defined as \cite{geng2026mean}:
\begin{equation}
    u(z_t,r,t)=\frac{1}{t-r}\int_{r}^{t} v(z_t,\tau)\,d\tau
\label{eq:5}
\end{equation}
where $z_t$ is the flow path and $z_t$= $a_t x+b_t \epsilon$, $a_t$ and $b_t$ are predefined schedules ($a_t=1-t$ and $b_t=t$ commonly). The instantaneous velocity $v_t=z_t'=a_t' x+b_t' \epsilon$, where $'$ denotes the time derivative and as $r \rightarrow t$, there is $\lim_{r \to t} u=v$. To have a formulation amenable to training, we rewrite Eq. \eqref{eq:5} as:
\begin{equation}
    (t-r)u(z_t,r,t)=\int_{r}^{t} v(z_t,\tau)\,d\tau.
\label{eq:6}
\end{equation}
We differentiate both sides with respect to $t$, treating $r$ as independent of $t$ and rearrange the terms, we could get:
\begin{equation}
    u(z_t,r,t)=v(z_t,t)-(t-r)\frac{d}{dt}u(z_t,r,t)
\label{eq:7}
\end{equation}
To compute the $\frac{d}{dt}u$ term in Eq. \eqref{eq:6}, note that $\frac{d}{dt}$ denotes a total derivative, which can be expanded in terms of partial derivatives:
\begin{equation}
    \frac{d}{dt}u(z_t,r,t)=\frac{dz_t}{dt}\partial_z u+\frac{dr}{dt}\partial_r u+\frac{dt}{dt}\partial_tu
    \label{eq:8}
\end{equation}
Eq. \eqref{eq:4} shows that $\frac{dz_t}{dt}=v(z_t,t)$ and $\frac{dr}{dt}=0$, $\frac{dt}{dt}=1$, we could get another relation between $u$ and $v$:
\begin{equation}
    \frac{d}{dt}u(z_t,r,t)=v(z_t,t)\partial_zu+\partial_tu
    \label{eq:9}
\end{equation}
The total derivative, as shown in this equation, is simply the Jacobian–vector product (JVP) of the Jacobian matrix $[\partial_z u,\partial_ru,\partial_tu]$ with the tangent vector $[v,0,1]$. Finally, we could get the target average velocity $u_{tgt}$ which serves as the ground-truth signal:
\begin{equation}
    u_{tgt}=v(z_t,t)-(t-r)(v(z_t,t)\partial_zu_{\theta}+\partial_t u_\theta)
    \label{eq:10}
\end{equation}
The velocity in Eq. \eqref{eq:9} is marginal velocity in Flow Matching \cite{lipman2022flow} and we replace it with conditional velocity:
\begin{equation}
    u_{tgt}=v_t-(t-r)(v_t\partial_zu_{\theta}+\partial_t u_\theta)
    \label{eq:11}
\end{equation}
In SAR-to-optical image translation, instantaneous velocity often focuses on local variations, potentially overlooking global structural coherence but average velocity accounts for the overall migration trend in the path integral, helping to preserve global contours and spatial layouts of ground objects during generation, thus reducing geometric distortions.

\subsection{Training with Contrastive Learning and Average Velocity}
To enable the model to distinguish optical images generated from different SAR images, thereby improving the quality of SAR‑to‑optical image translation, We adopt contrastive learning to make the flows predicted from different SAR images distinct from each other. In detail, Given a batch $B$ of SAR images $s_0,s_1...s_{B-1}$ and their paired optical images $x_0,x_1...x_{B-1}$, we first construct the interpolation path $z_t=(1-t)x_0+t\epsilon$, where $\epsilon \sim N(0,I)$ and then compute the instantaneous velocity $v_t=\epsilon-x_0$. Next, for each sample in the batch, we concatenate the interpolation path$z_t$ with the each SAR image in the batch $B$ along the channel dimension and get the concatenated tensors $c_0,c_1...c_{B-1}$. The concatenated tensors $c_i$ are then passed through a VAE encoder (with frozen parameters) for spatial dimensionality reduction. The resulting compressed feature maps are fed into a U‑Net architecture, which comprises ResNet blocks and self‑attention blocks. Subsequently, we apply a Jacobian‑Vector Product (JVP) operation to the U‑Net output to obtain the predicted average velocities $u_0,u_1......u_{B-1}$ and the partial time derivative$\partial_{z_0}u^0_\theta+\partial_{t}u^0_\theta+,...,\partial_{z_{B-1}}u^{B-1}_\theta+\partial_{t}u^{B-1}_\theta$. The target average velocity $u_{tgt}^0,u_{tgt}^1,...,u_{tgt}^{B-1}$ is then computed according to Eq. \eqref{eq:10}. 
\begin{equation}
    u_{tgt}^i=v(z_t,t)-(t-r)(v(z_t,t)\partial_{z_i}u^i_{\theta}+\partial_t u^i_{\theta})
    \label{eq:12}
\end{equation}
where $i$ ranges from $0$ to $B-1$. Finally, the loss function is defined as follows:
\begin{equation}
  \mathcal{L} = \|u_0 - u_{tgt}^0\|^2 - \frac{\lambda}{B-1} \sum_{i=1}^{B-1} \|u_0 - u_{tgt}^i\|^2 
  \label{eq:13}
\end{equation}
where $\lambda$ is a fixed parameter that controls the strength of contrastive regularization.

\subsection{One-step Inference}
Conventional GAN-based methods and diffusion-based methods could generate high-quality optical images but they face challenge that the inference latency is high, to reduce the inference latency, we design the one-step inference method. Given $\epsilon \sim p_{prior}(\epsilon)$, r=0, t=1 and for one-step SAR-to-optical translation, we have:
\begin{equation}
    x=z_r=z_t-(t-r)u(z_t,0,1) \rightarrow x= \epsilon-u(z_1,0,1)
    \label{eq:14}
\end{equation}

\section{Experiments}
\label{sec:experiments}

\subsection{Datasets and Evaluation Metrics}
To comprehensively assess the generalization capability of our model for SAR‑to‑optical translation under varying scene types and seasonal conditions, we utilize two datasets: QXS-SAROPT \cite{huang2021qxs} and SAR2Opt \cite{zhao2022comparative}. The QXS-SAROPT dataset contains 20,000 pairs of SAR and optical images collected from the Gaofen-3 satellite and Google Earth, mainly covering several port cities. Each image patch has a size of 256 by 256. The approximate split ratio of training and test set for QXS-SAROPT is 4:1. The SAR2Opt dataset provides 2,076 pairs of SAR and optical images acquired from the TerraSAR-X satellite and Google Earth, covering multiple Asian cities. Each image patch has a size of 600 by 600. This dataset follows the official split, with 1,450 images in the training set and 627 images in the test set. The evaluation metrics we use are SSIM \cite{hore2010image}, FID \cite{yu2021frechet}, PSNR \cite{hore2010image}, and LPIPS \cite{zhang2018unreasonable}. S:SSIM, F:FID, L:LPIPS, P: PSNR 
\begin{figure}[htbp]
    \centering
    \begin{subfigure}[b]{0.1\textwidth}
        \centering
        \includegraphics[width=\textwidth]{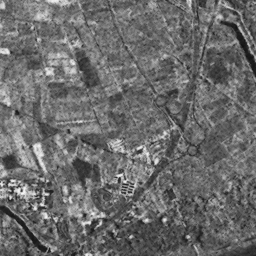}
        \subcaption*{}
        \label{fig:sample1_sar}
    \end{subfigure}
    \begin{subfigure}[b]{0.1\textwidth}
        \centering
        \includegraphics[width=\textwidth]{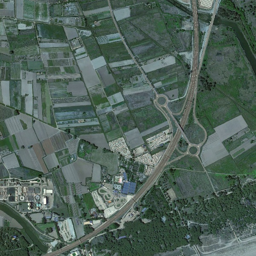}
        \subcaption*{}
        \label{fig:sample1_gt}
    \end{subfigure}
    \begin{subfigure}[b]{0.1\textwidth}
        \centering
        \includegraphics[width=\textwidth]{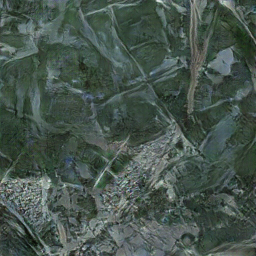}
        \subcaption*{}
        \label{fig:sample1_pcgl}
    \end{subfigure}
    \begin{subfigure}[b]{0.1\textwidth}
        \centering
        \includegraphics[width=\textwidth]{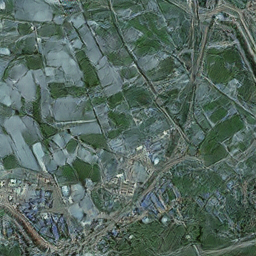}
        \subcaption*{}
        \label{fig:sample1_ours}
    \end{subfigure}


    \begin{subfigure}[b]{0.1\textwidth}
        \centering
        \includegraphics[width=\textwidth]{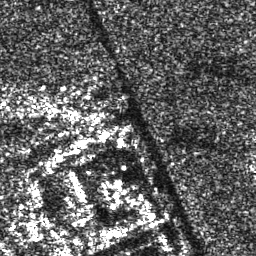}
        \caption{SAR}
        \label{fig:sample2_sar}
    \end{subfigure}
    \begin{subfigure}[b]{0.1\textwidth}
        \centering
        \includegraphics[width=\textwidth]{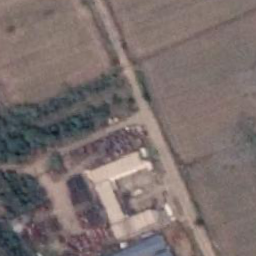}
        \caption{GT}
        \label{fig:sample2_gt}
    \end{subfigure}
    \begin{subfigure}[b]{0.1\textwidth}
        \centering
        \includegraphics[width=\textwidth]{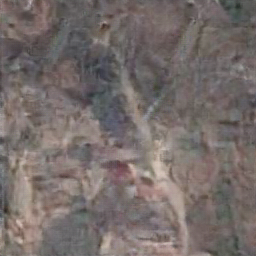}
        \caption{PCGL-GAN}
        \label{fig:sample2_pcgl}
    \end{subfigure}
    \begin{subfigure}[b]{0.1\textwidth}
        \centering
        \includegraphics[width=\textwidth]{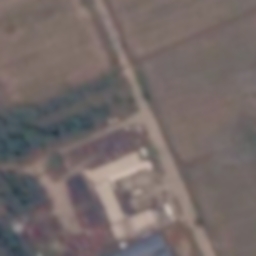}
        \caption{Ours}
        \label{fig:sample2_ours}
    \end{subfigure}

    \caption{Comparison of SAR-to-opticacl image translation results between the second-best model PCGL-GAN and our model. First row: SAR2Opt test set. Second row: QXS-SAROPT test set. }
    \label{fig:comparison}
\end{figure}

\subsection{Baselines}
The baselines we use are divided into GAN-based methods and diffusion-based methods. GAN-based methods include Pix2Pix \cite{pix2pix2017}, CycleGAN \cite{CycleGAN2017}, and PCGL-GAN \cite{chen2026towards}. Diffusion-based methods include S2ODPM \cite{bai2023conditional}, cDDPM \cite{kim2025conditional} and CycleDiff \cite{zou2026cyclediff}. Contrastive learning-based methods include KANCUT \cite{mahara2025dawn} and CUT \cite{park2020contrastive}.
\subsection{Implementation Details}
We use 8×NVIDIA A100 PCIE (80GB) for training and a single NVIDIA A100 PCIE (80GB) for inference. We train our ContraFM-S2O for 400 epochs with the learning rate of 2e-4 and batch size of 128, the $\lambda$ we set in Eq. \eqref{eq:13} is 0.01.
\subsection{Main Results}
To evaluate our model, we train and test it on the QXS and SAR2Opt datasets, using SSIM, FID, LPIPS, and PSNR as evaluation metrics. We also compare the average inference latency per image between ContraFM‑S2O and the baseline models. The inference steps for all diffusion-based models are 100, and our ContraFM-S2O employs one-step inference.
\begin{table}[htbp]
\centering
\caption{Performance comparison of different baseline models on two datasets. The best results are highlighted in \textbf{bold} and the second-best results are \underline{underlined} }
\label{tab:comparison_results}
\footnotesize
\setlength{\tabcolsep}{2.3pt}
\begin{tabular}{lccccccccccc}
\toprule
\multirow{2}{*}{Model} & \multicolumn{4}{c}{QXS} & \multicolumn{4}{c}{SAR2Opt} & \multirow   {2}{*}{Latency}\\
\cmidrule(lr){2-5} \cmidrule(lr){6-9}
 & S & F & L &P & S & F & L &P  \\
\midrule
Pix2Pix & 0.235 & 130.2 & 0.451 & 23.12 & 0.138 & 166.1 & 0.384 & 26.12 & 0.43s \\
CycleGAN  & 0.241 & 126.8 & 0.438 & 27.92 & 0.140 & 153.5 & 0.341 & 27.98 & \underline{0.38s}  \\
PCGL-GAN & \underline{0.279} & \underline{107.2} & 0.349 & \underline{28.02} & \underline{0.175} & \underline{144.4} & \underline{0.316} & \underline{28.18} & 0.44s  \\
S2ODPM & 0.262 & 130.3 & 0.392 & 21.25 & 0.152 & 148.3 & 0.338 & 28.02 & 1.22s  \\
cDDBM & 0.253 & 141.2 & 0.414 & 19.24 & 0.144 & 157.7 & 0.397 & 27.69 & 1.53s  \\
CycleDiff & 0.241 & 135.7 & 0.376 & 27.92 & 0.149 & 165.6 & 0.379 & 28.04 & 1.69s  \\
CUT & 0.236 & 113.5 & 0.383 & 27.95 & 0.126 & 162.4 & 0.359 & 27.99 & 0.83s    \\
KANCUT & 0.229 & 119.7 & \underline{0.341} & 27.96 & 0.123 & 177.8 & 0.373 & 27.92 & 0.98s \\
\textbf{Ours} & \textbf{0.298} & \textbf{102.42} & \textbf{0.311} & \textbf{31.22} & \textbf{0.196} & \textbf{137.6} & \textbf{0.307} & \textbf{29.03} & \textbf{0.10s} \\
\bottomrule
\end{tabular}
\label{main_results}
\end{table}
Table. \ref{main_results} shows that our ContraFM-S2O achieves the state-of-the-art (sota) performance on QXS-SAROPT and SAR2Opt datasets across baselines, which demonstrates the effectiveness of ContraFM-S2O utilizing contrastive learning and modeling the average velocity. In addition, ContraFM-S2O could achieve \textbf{4.4×} inference speed acceleration compared to PCGL-GAN, which achieves the second-best SAR-to-optical image translation results. See Fig \ref{fig:comparison} for the qualitative results.
\subsection{Ablation Studies}
We conduct ablation studies on the influence of average velocity and contrastive learning, as well as the influence of $\lambda$ in Eq. \eqref{eq:13}. We employ one-step inference in all ablations studies.
\subsubsection{The Influence of Average Velocity and Contrastive Learning}

\begin{table}[htbp]
\centering
\caption{The influence of average velocity and contrastive learning. We compare ContraFM-S2O with variants without average velocity, without contrastive learning and without both, w/o: without. The best results are highlighted in \textbf{bold} and the second-best results are \underline{underlined}}
\label{tab:ablation2}
\footnotesize
\setlength{\tabcolsep}{2pt}
\begin{tabular}{lcccccccccc}
\toprule
\multirow{2}{*}{Model} & \multicolumn{4}{c}{QXS} & \multicolumn{4}{c}{SAR2Opt} & \multirow   {2}{*}{Latency} \\
\cmidrule(lr){2-5} \cmidrule(lr){6-9}
 & S & F & L & P & S & F & L & P \\
\midrule
Full & \textbf{0.298} & \textbf{102.42} & \textbf{0.311} & \textbf{31.22} & \textbf{0.196} & \textbf{137.6} & \textbf{0.307} & \textbf{29.03} & \textbf{0.10s} \\
w/o Avg & \underline{0.243} & \underline{158.91} & \underline{0.357} & \underline{27.77} & 0.134 & \underline{167.5} & \underline{0.373} & \underline{25.33} & 0.52s  \\
w/o CL & 0.197 & 184.09 & 0.407 & 23.24 & \underline{0.183} & 210.9 & 0.416 & 22.55 & \underline{0.14s}  \\
w/o Both & 0.148 & 235.96 & 0.454 & 19.17 & 0.117 & 254.3 & 0.465 & 19.87 & 0.73s  \\
\bottomrule
\end{tabular}
\end{table}

\noindent Average velocity primarily speeds up inference (via simplifying integration to vector subtraction), while contrastive learning improves image quality by preventing velocity‑field overlap, enabling detailed edge generation (Table \ref{tab:ablation2}). Optimal quality and speed occur at $\lambda=0.01$; larger $\lambda$ degrades quality by weakening SAR‑optical feature association (Table \ref{tab:ablation3}).
\subsubsection{The Influence of $\lambda$ during training}
\begin{table}[htbp]
\centering
\caption{The influence of different $\lambda$ during training. The best results are highlighted in \textbf{bold} and the second-best results are \underline{underlined}}
\label{tab:ablation3}
\footnotesize
\setlength{\tabcolsep}{2pt}
\begin{tabular}{lcccccccccc}
\toprule
\multirow{2}{*}{$\lambda$} & \multicolumn{4}{c}{QXS} & \multicolumn{4}{c}{SAR2Opt} & \multirow   {2}{*}{Latency} \\
\cmidrule(lr){2-5} \cmidrule(lr){6-9}
 & S & F & L & P & S & F & L & P \\
\midrule
0 & 0.197 & 184.09 & 0.407 & 23.24 & \underline{0.183} & 210.9 & 0.416 & 22.55 & 0.14s \\
0.01 & 0.298 & \textbf{102.42} & \textbf{0.311} & \textbf{31.22} & \textbf{0.196} & \textbf{137.6} & \textbf{0.307} & \textbf{29.03} & \textbf{0.10s}  \\
0.05 & \textbf{0.304} & 113.44 & 0.333 & 26.71 & 0.191 & 176.7 & 0.388 & 26.55 & \underline{0.13s}  \\
0.1 & 0.154 & 167.29 & 0.389 & 22.35 & 0.143 & 189.1 & 0.424 & 21.65 & 0.29s  \\
0.2 & 0.127 & 209.23 & 0.411 & 20.87 & 0.128 & 222.1 & 0.477 & 18.38 & 0.28s  \\
\bottomrule
\end{tabular}
\end{table}

\section{Conclusion}
\label{sec:conclusion}
We propose ContraFM-S2O, a flow-matching model for SAR-to-optical translation. It improves generation quality via contrastive learning and speeds up inference via average velocity modeling. Experimental results demonstrate that our model consistently achieves state‑of‑the‑art performance across multiple benchmark datasets, outperforming various baseline models in terms of both inference speed and generation quality, validating that contrastive learning enhances input discrimination and detail preservation, while average velocity reduces computation by replacing multi-step ODE integration with a single efficient vector operation—together achieving strong speed-quality synergy. 
\vfill\pagebreak

\label{sec:refs}
\bibliographystyle{IEEEbib}
\bibliography{refs}

\end{document}